\documentclass[]{TEAI}
\usepackage{array}

\newcommand{\method}{IDSpect}
\newcommand{\lev}{\operatorname{Lev}}

\title{Decompose Radicals, Then Reward: Fine-Grained Inspection for Accurate Chinese Text Rendering}

\author[1,2,*]{Yazhen Xie}
\author[1,2,*]{Xingsong Ye}
\author[1,2,\dagger]{Zhineng Chen}

\affiliation[1]{\mbox{Institute of Trustworthy Embodied AI, Fudan University}}
\affiliation[2]{\mbox{Shanghai Key Laboratory of Multimodal Embodied AI}}

\contribution[*]{Equal contribution}
\contribution[\dagger]{Corresponding author}

\abstract{
Rendering accurate Chinese text remains challenging for text-to-image models.
Existing OCR-based reinforcement-learning rewards compare decoded transcripts
with target strings. Such rewards overlook the compositional nature of Chinese
writing: an ideograph consists of reusable components arranged through explicit
spatial relations, yet OCR evaluates it as an atomic character. Consequently,
visually different radical-level errors may receive equally
coarse feedback, encouraging glyphs that merely resemble the target instead of
faithfully reproducing its internal structure. We employ
Ideographic Description Sequences (IDS), which comprise spatial operators and
character components, and train an expert IDS recognizer to
transcribe rendered Chinese text into this representation. Building on this
recognizer, we introduce \textbf{\method}, which deterministically decomposes the
target text into IDS tokens and aligns crop-level visual IDS predictions with
the target sequence. Globally unique token credit makes this comparison robust
to the order of detected text regions. Combined with a whole-character semantic
reward, \method{} supplies fine-grained credit with component and spatial-relation without
changing the image generator or adding inference-time cost. Experiments with
GRPO post-training of Qwen-Image demonstrate that \method{} achieves leading
structural quality and semantic alignment on LongText and GenTextEval.
}

\checkdata[Keywords]{Text-to-Image Generation, Chinese Visual Text Rendering, OCR Reward}

\correspondence{\email{zhinchen@fudan.edu.cn}}

\begin{document}
\maketitle

\section{Introduction}
\label{sec:intro}

\begin{figure}[t]
  \centering
  \includegraphics[width=0.7\textwidth]{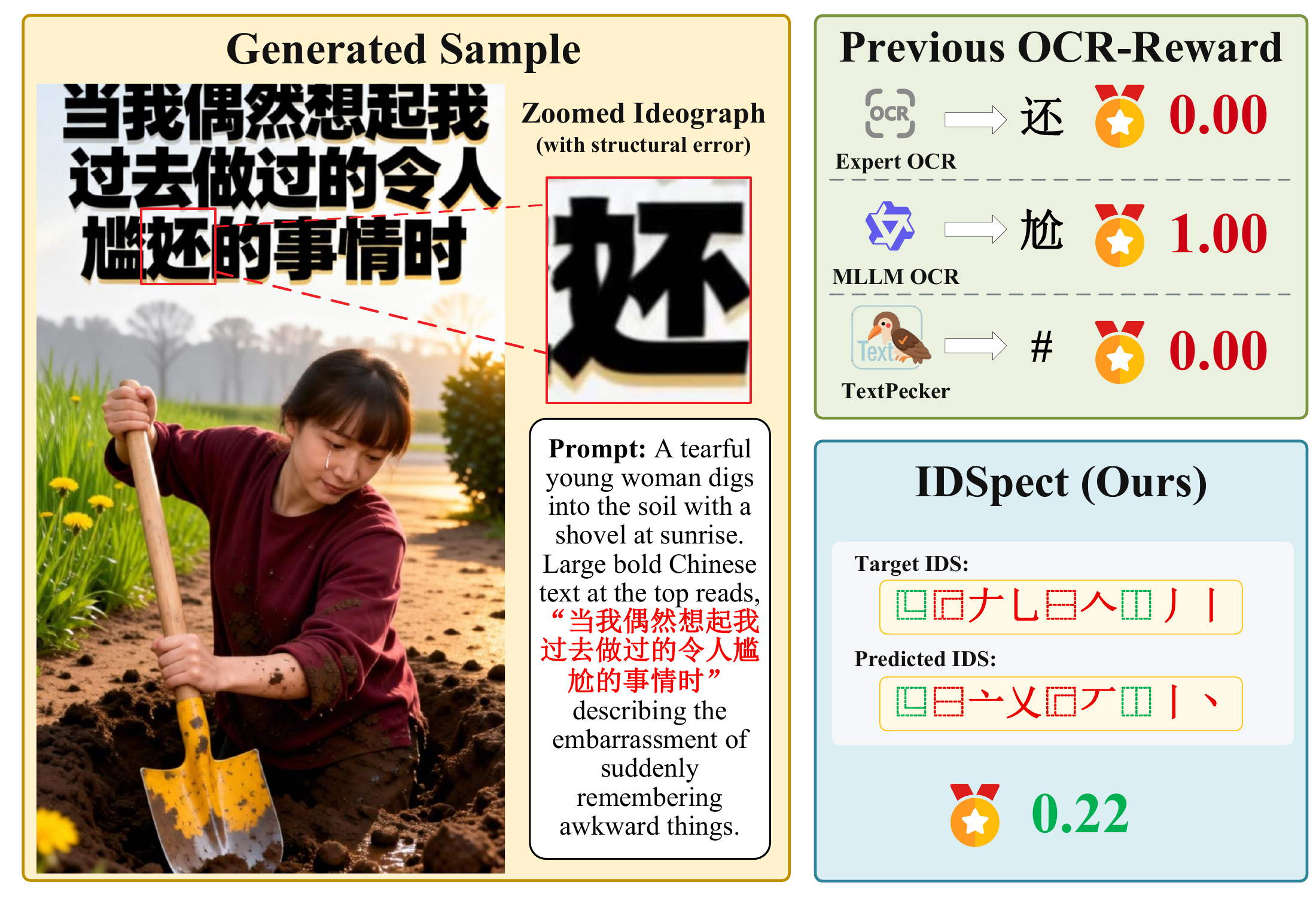}
  \caption{Local reward responses to a malformed ideograph. Character-level
  rewards produce discrete decisions, whereas \method{} retains partial
  structural credit via IDS matching.
  }
  \label{fig:motivation}
\end{figure}

Text-to-image generation has advanced rapidly~\cite{rombach2022high,
QwenImage}. Specialized encoders and explicit glyph
conditioning have further improved visual text
rendering~\cite{tuo2024anytext,chen2023textdiffuser, textssr}.
Nevertheless, rendering specified text inside an image
remains a persistent failure mode, particularly for Chinese
ideographs~\cite{zhang2025ocrgenbench}.
Unlike alphabetic characters, a Chinese ideograph can encode a two-dimensional
composition of reusable components under left--right, top--bottom, enclosing,
and other spatial relations~\cite{unicode16}.
This large and long-tailed output space makes both generation and evaluation
difficult: a glyph may preserve several correct components while corrupting
only one component or their spatial arrangement.

Reinforcement-learning post-training offers a direct way to optimize text
rendering~\cite{liu2025flowgrpo}, and OCR similarity is a natural reward for
this objective~\cite{TextPecker}.
However, an OCR reward first compresses the image into a character transcript
and then evaluates that transcript. Consequently, it provides weak credit for
partial structural progress: repairing one radical of a still-misrecognized
ideograph may leave the reward unchanged. Conversely, linguistic priors can
occasionally recover the intended character from a malformed glyph, producing
a high semantic score without adequate visual evidence.

Fig.~\ref{fig:motivation}
shows both failure modes in the same generated image. For the highlighted
target ideograph, a specialist OCR model follows the local appearance and
predicts a visually plausible but incorrect character, contributing 0.00 to
the character-level reward. A global MLLM OCR instead recovers the target from
the surrounding linguistic context despite its malformed structure, producing
a false-positive local contribution of 1.00. TextPecker detects or quantifies anomalous glyphs and is representative of
recent structure-aware evaluators for visual text
rendering~\cite{TextPecker}.
TextPecker correctly identifies the same glyph as
structurally anomalous and replaces it with \texttt{\#}, avoiding the MLLM's
false positive but assigning a local contribution of 0.00. Once the glyph is
collapsed into an anomaly marker, however, its correct and incorrect internal
structures are no longer distinguished.

\begin{figure}[tbp]
  \centering
  \includegraphics[width=\textwidth]{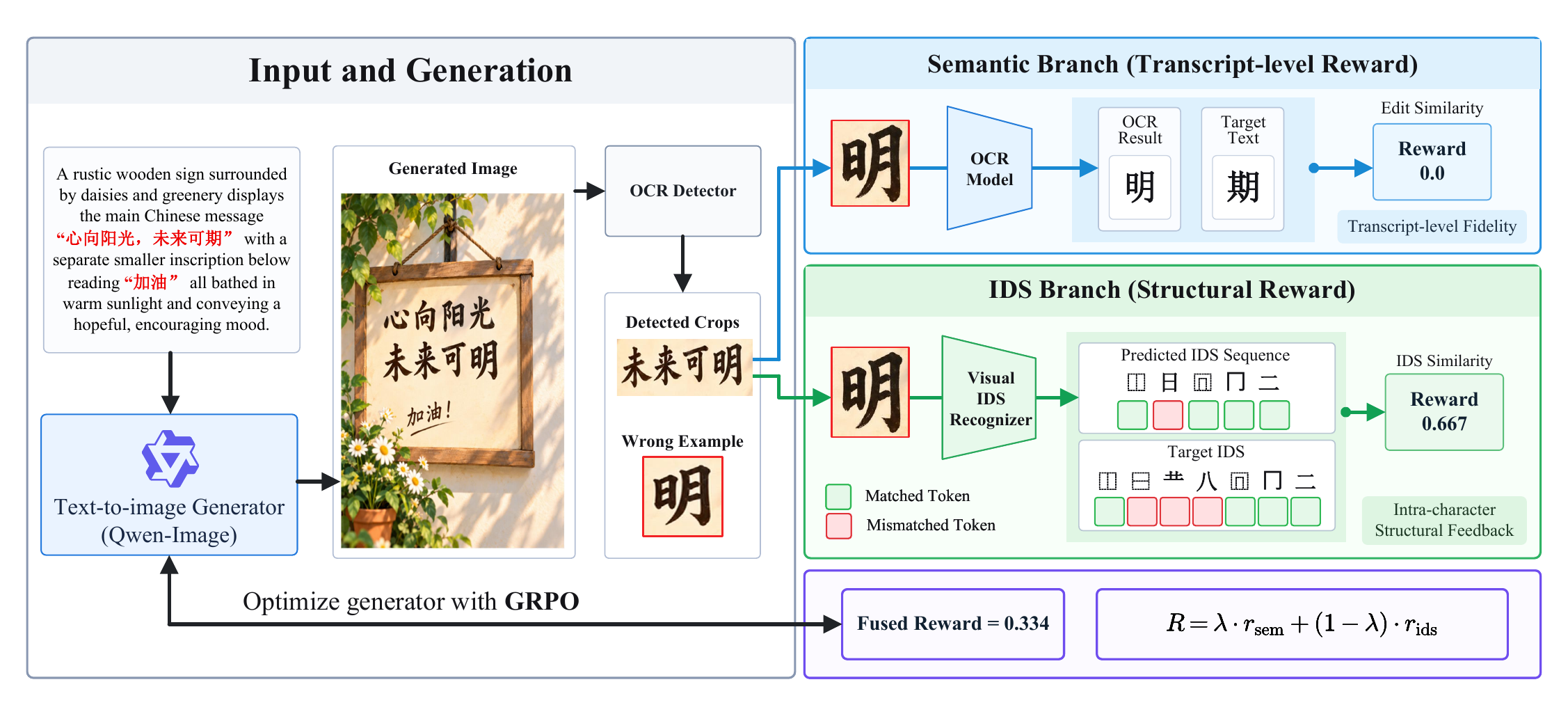}
  \caption{Framework of \method. An OCR detector extracts text crops from each
  generated image. The semantic branch compares OCR transcripts with the
  target text, while the IDS branch compares visually predicted and target IDS
  tokens for intra-character feedback. Their weighted sum guides GRPO
  post-training of Qwen-Image.}
  \label{fig:overview}
\end{figure}

Prior Chinese OCR work uses radicals, strokes, and IDS to recognize rare or
unseen characters~\cite{Yu_2023_ICCV_CCRCLIP,
zhang2018radical,ijcai2021p0085}.
We build on these representations to assess how much of a generated
ideograph's internal structure matches its target, providing partial
credit even when the whole character is incorrect.
To this end, we introduce \textbf{\method}, illustrated in
Fig.~\ref{fig:overview}. We first train an expert IDS recognizer that transcribes
detected text regions into IDS token sequences. During post-training, the
detected crops feed two complementary reward branches: the semantic branch
compares OCR transcripts with the target text, while the structural branch
aligns visual IDS predictions with the deterministic target IDS. Their weighted
sum combines transcript-level fidelity with intra-character structural credit.
In Fig.~\ref{fig:motivation}, \method{} retains credit for matched IDS tokens
while penalizing mismatched ones, assigning a standalone IDS reward of 0.22.
Thus, anomaly-aware rewards identify which glyph is problematic, whereas a
compositional reward additionally quantifies how much of its internal structure
already matches the target.

Our contributions are threefold:
\begin{itemize}
  \item We identify the coarse-credit limitation of atomic OCR rewards for
  Chinese visual text generation, under which structurally distinct
  intra-character errors can receive indistinguishable feedback.
  \item We develop an expert IDS recognizer that decomposes
  rendered text in generated images into sequences of spatial operators and
  character components.
  \item We construct \method{} as a fine-grained
  reward for GRPO. On LongText and GenTextEval, it improves both structural
  quality and semantic alignment over general OCR-based rewards and TextPecker.
\end{itemize}

\section{Related Work}
\label{sec:related}

\subsection{Visual text generation and post-training}
Text-to-image models have progressively improved spelling and layout through specialized encoders and explicit glyph conditions~\cite{tuo2024anytext,tuo2024anytext2,Glyphdraw2,chen2023textdiffuser,chen2024textdiffuser2,QwenImage}. More broadly, recent diffusion-based generation methods have explored fine-grained visual conditioning and text-driven control to achieve more precise correspondence between visual prompts and generated content~\cite{yang20233dstyle,yang2024hi3d}. Recent methods introduce hierarchical rewards and region-level preference optimization for visual text rendering~\cite{cui2026textalign,Shuai_2026_CVPR_GlyphPrinter}.
When used as rewards, specialist OCR models~\cite{du2025svtrv2,ye2026advancing,li2026hunyuanocr,ye2026all} such as
PP-OCRv5~\cite{PPOCRV5} directly optimize transcription accuracy but inherit the
recognizer's atomic label space. TextPecker~\cite{TextPecker} introduces
structural anomaly quantification for reward-guided generation. However, it roughly recognizes all incorrectly written characters as ``\#'', which is overly coarse. This motivates us to measure edit progress over the internal composition of a target ideograph rather than only assigning a quality judgment to the whole glyph or text region.

\subsection{Structure-aware Chinese text recognition}
Radicals, strokes, and IDS have long been used to improve recognition of rare
or unseen Chinese characters~\cite{Yu_2023_ICCV_CCRCLIP,zhu2025zero,
zhang2018radical,ijcai2021p0085}.
These studies use decomposition to recognize text. Inspired by them, we instead train an IDS reader as a reward model and use its structured output to optimize a generative model. Reward utility depends not only on final recognition accuracy, but also on whether intermediate scores rank partially correct and malformed glyphs in a useful order.

\section{Method}
\label{sec:method}

\subsection{Problem formulation}

Let $p$ be an image description and $y=(c_1,\ldots,c_M)$ the text that should be visible in the generated image. A generator $G_\theta$ samples $I\sim G_\theta(p,y)$. During group-relative post-training, multiple images are sampled under the same condition and scored by a reward $R(I,y)$~\cite{shao2024deepseekmath,liu2025flowgrpo}. Our goal is to construct a reward that preserves semantic correctness while providing fine-grained supervision for the internal composition of Chinese ideographs.

An OCR model $f_{\mathrm{ocr}}$ is applied to the generated image to obtain the recognized transcript $\hat y$. Let $\bar y$ and $\bar{\hat y}$ denote the target and recognized strings after removing spaces and lowercasing. We measure semantic fidelity using normalized character edit similarity:
\begin{equation}
 r_{\mathrm{sem}}(I,y)
 = \begin{cases}
  1, & \bar y\text{ occurs in }\bar{\hat y},\\
  1-\dfrac{\min\{\lev(\bar{\hat y},\bar y),|\bar y|\}}{|\bar y|},
     & \text{otherwise}.
 \end{cases}
 \label{eq:semantic}
\end{equation}
All training targets are nonempty, and the truncated edit distance keeps the reward within $[0,1]$. The substring condition gives full credit when the target text is correctly recognized even if the OCR output contains additional characters. However, this semantic reward treats each character as an atomic symbol and therefore provides limited information about partially correct character structures. We address this limitation by introducing an IDS-based compositional reward in the following section.

\subsection{Balanced IDS recognizer}

IDS represents a Chinese ideograph as a hierarchical tree, where internal nodes correspond to spatial operators and leaf nodes correspond to reusable character components~\cite{unicode16}. In this representation, a normal Chinese character can be recursively parsed into a sequence of structural operations and primitive components, providing an explicit description of how the character is spatially composed. For example, a character formed by placing two components side by side is represented by a horizontal composition operator together with the IDS representations of its two subcomponents. The decomposition can then be recursively applied to each subcomponent until reaching indivisible or reusable components. In this way, IDS converts the implicit two-dimensional composition of a Chinese character into an explicit hierarchical structure that can be serialized as a token sequence.

Using the Unicode 16.0 BabelStone lexicon~\cite{unicode16,babelstone_ids}, we recursively decompose the Chinese characters covered by GB~18030--2022~\cite{gb18030_2022}. Let $D(c)$ denote the flattened IDS sequence obtained by recursively decomposing a character $c$. For a text string $y=(c_1,\ldots,c_M)$, its target structural representation is constructed by concatenating the IDS sequences of all characters, i.e., $D(y)=D(c_1)\oplus\cdots\oplus D(c_M)$. We cap the recursive decomposition depth at 10 and construct the decoder vocabulary from the resulting spatial operators and reusable components.

Based on this representation, the IDS recognizer $f_{\mathrm{ids}}$ is trained to directly predict the IDS sequence from a rendered text image rather than first recognizing the image as a character sequence and then performing symbolic decomposition. Given an input text image $x$, the recognizer produces $\hat{D}=f_{\mathrm{ids}}(x)$, where $\hat{D}$ is a sequence of spatial operators and character components. Thus, the recognition process transforms the visual appearance of the text directly into its underlying structural representation. This formulation shifts the recognition target from a flat character identity to the internal composition of each ideograph, enabling the model to explicitly recover the structural primitives and their spatial relationships from visual evidence.

Specifically, we train an IDS recognizer $f_{\mathrm{ids}}$ that maps a text-region
image $I_k$ to an IDS sequence $\hat s_k$. Building on the SVTRv2 visual
backbone~\cite{du2025svtrv2}, a recent state-of-the-art architecture for scene
text recognition, we reformulate character transcription as autoregressive
prediction over the IDS vocabulary. The resulting recognizer couples an
SVTRv2 encoder with a Transformer-style decoder following
NRTR~\cite{sheng2019nrtr} to directly recover the compositional structure of
rendered text.

Crucially, training data determine whether this recognizer can reliably parse
the long tail of Chinese ideographs. Natural scene-text corpora are strongly
biased toward frequent characters, and their IDS annotations consequently
provide highly imbalanced supervision for radicals and other components. We
therefore adapt the UnionST rendering engine~\cite{ye2026wrong_UnionST} to
construct a balanced synthetic training set (IDSynth-1M).
We form a rare-ideograph-enriched corpus with balanced frequencies
across covered characters, render its strings into text
images, and pair each image with its IDS sequence. This design
does more than increase rare-character coverage. With naturally distributed
text, the recognizer can exploit character-frequency and linguistic shortcuts:
it first recognizes a glyph as a frequent character and then reproduces its
canonical IDS, without grounding the output in the observed
components. Such behavior reduces IDS prediction to character recognition and
cannot expose internal glyph errors. Balanced, randomly composed transcripts
discourage this shortcut and promote component-grounded prediction.

\subsection{Target-conditioned compositional reward}

The OCR detector yields text crops $\{I_k\}_{k=1}^{K}$, and the IDS recognizer produces a structural token sequence $\hat{s}_k=f_{\mathrm{ids}}(I_k)$ for each crop. Let $s=D(y)$ denote the complete IDS sequence of the target text. Rather than concatenating the predicted sequences in detector order, we align each $\hat{s}_k$ with candidate spans of $s$, since detected crops may appear in an arbitrary order and may cover different portions of the target text. Candidate spans are generated by semi-global edit alignment and filtered by non-maximum suppression to remove near-duplicate matches. A global assignment then selects at most one target span for each predicted crop, while each target position can receive exact-match credit only once. This makes the matching independent of crop order while preserving the IDS-token order within each crop and its aligned target span.

Let $A^\star$ denote the selected global assignment and $C(A^\star)$ the number of uniquely credited target IDS tokens. We define the compositional reward as a token-level F1 score
\begin{equation}
 r_{\mathrm{ids}}(I,y)
 =\frac{2C(A^\star)}
 {|D(y)|+\sum_{k=1}^{K}|\hat{s}_k|}.
 \label{eq:ids_target}
\end{equation}
The reward provides fine-grained structural supervision because matching individual components and spatial operators can increase the score even when the atomic OCR output remains unchanged. At the same time, missing or extraneous predictions are penalized through the F1 denominator and the unique-credit constraint. Finally, we combine the semantic and compositional rewards as
\begin{equation}
 R(I,y)=\lambda r_{\mathrm{sem}}(I,y)
 +(1-\lambda)r_{\mathrm{ids}}(I,y),
 \label{eq:combined}
\end{equation}
where $r_{\mathrm{sem}}$ provides target-level semantic supervision and $r_{\mathrm{ids}}$ provides fine-grained structural supervision. The fused reward is used to guide GRPO post-training of $G_\theta$. Both reward branches are discarded after training, so \method{} introduces no additional cost during generator inference.

\begin{table}[tbp]
\small
\centering
\caption{Quantitative comparison of Qwen-Image variants on Chinese visual text
rendering benchmarks. Base denotes the frozen model. OCR,
TextPecker~\cite{TextPecker}, and \method{} denote GRPO post-training with the
corresponding rewards. \textbf{Avg.}: original benchmark text score,
\textbf{Qua.}: structural quality, \textbf{Sem.}: semantic alignment.
Qua. and Sem. are evaluated by TextPecker.}
\label{tab:main}
\setlength{\tabcolsep}{9mm}
\begin{tabular}{l|ccc|cc}
\toprule
\multirow{2}{*}{Rewards}
& \multicolumn{3}{c|}{LongText}
& \multicolumn{2}{c}{GenTextEval} \\
\cmidrule(lr){2-4}\cmidrule(lr){5-6}
& Avg. & Qua. & Sem. & Qua. & Sem. \\
\midrule
Base         & 0.920 & 0.924 & 0.834 & 0.933 & 0.810 \\
OCR        & 0.967 & 0.956 & 0.886 & 0.953 & 0.874 \\
TextPecker & \textbf{0.974} & 0.969 & 0.908
             & 0.973 & 0.897 \\
\method{}  & 0.972 & \textbf{0.975} & \textbf{0.928}
             & \textbf{0.979} & \textbf{0.911} \\
\bottomrule
\end{tabular}
\end{table}

\section{Experiments}
\label{sec:experiments}

\subsection{Experimental setup}

\textbf{Generator and optimization.}
We post-train Qwen-Image~\cite{QwenImage} with
Flow-GRPO~\cite{liu2025flowgrpo} using the Flow-Factory
framework~\cite{ping2026flowfactory}. Following its default Qwen-Image
configuration, we optimize LoRA adapters~\cite{hu2022lora} with rank $r=64$ and
scaling factor $\alpha=128$ using AdamW, with a learning rate of
$3\times10^{-4}$ and weight decay of $10^{-4}$. All remaining hyperparameters
follow the framework defaults.
For \method, we set $\lambda=0.5$, assigning equal weights to the semantic and
IDS rewards throughout all experiments.

\textbf{IDS-recognizer data.}
We construct \textbf{IDSynth-1M}, a dataset of one million rendered scene text images, to train \(f_{\mathrm{ids}}\). To mitigate the long-tailed distribution of natural Chinese text, we sample characters approximately uniformly from the entire character set covered by the selected Chinese fonts inherited from UnionST to form transcripts. This sampling strategy also suppresses natural lexical co-occurrence, making it difficult for the model to infer complete characters from linguistic context alone and thereby encouraging it to parse the IDS sequence primarily from visual information without interference from language modeling. We then convert each sampled transcript into its corresponding IDS representation as the training target. The maximum decoded IDS sequence length is set to 100 tokens.

\textbf{Baselines and metrics.}
Baselines include the frozen model, OCR-only GRPO, and TextPecker-guided
GRPO~\cite{TextPecker}. We compare them with \method{}-guided GRPO.
Following TextPecker's Chinese evaluation protocol, we report results on
LongText-Bench~\cite{X-Omni_RL_and_LongText} and
GenTextEval-Bench~\cite{TextPecker} using the original benchmark text score
(Avg.), structural quality (Qua.), and semantic alignment (Sem.). The published
Qwen-Image, OCR-reward, and TextPecker-reward results are included for direct
comparison. \method{} is evaluated using the same benchmark protocol.

\subsection{Experimental Results}

\textbf{Quantitative results.}
Tab.~\ref{tab:main} summarizes the quantitative comparison on LongText and GenTextEval. Relative to the frozen Qwen-Image model, OCR-guided GRPO substantially improves the original benchmark text score as well as both TextPecker-based structural quality and semantic alignment. On LongText, the original text score increases from 0.920 to 0.967, while Qua. and Sem. improve from 0.924 and 0.834 to 0.956 and 0.886, respectively. Replacing the OCR reward with the TextPecker reward brings further improvements, achieving an Avg. score of 0.974 on LongText together with Qua. and Sem. scores of 0.969 and 0.908. On GenTextEval, TextPecker-guided GRPO reaches 0.973 Qua. and 0.897 Sem. Our \method{} further improves both structural and semantic metrics, achieving 0.979 Qua. and 0.911 Sem. on GenTextEval. These results exceed the OCR reward by 0.026 and 0.037, respectively, and improve over the TextPecker reward by 0.006 and 0.014. On LongText, \method{} achieves the highest Qua. and Sem. scores of 0.975 and 0.928, respectively, while maintaining an Avg. score of 0.972 that is comparable to TextPecker. These results indicate that IDS-based supervision provides a finer-grained signal for Chinese character rendering. Rather than treating each character as a single categorical recognition target, the IDS reward evaluates whether its internal components and spatial composition are correctly rendered, which is particularly beneficial for correcting localized structural errors that may be overlooked by conventional OCR-based rewards.

\begin{figure}[!t]
  \centering
  \includegraphics[width=\textwidth]{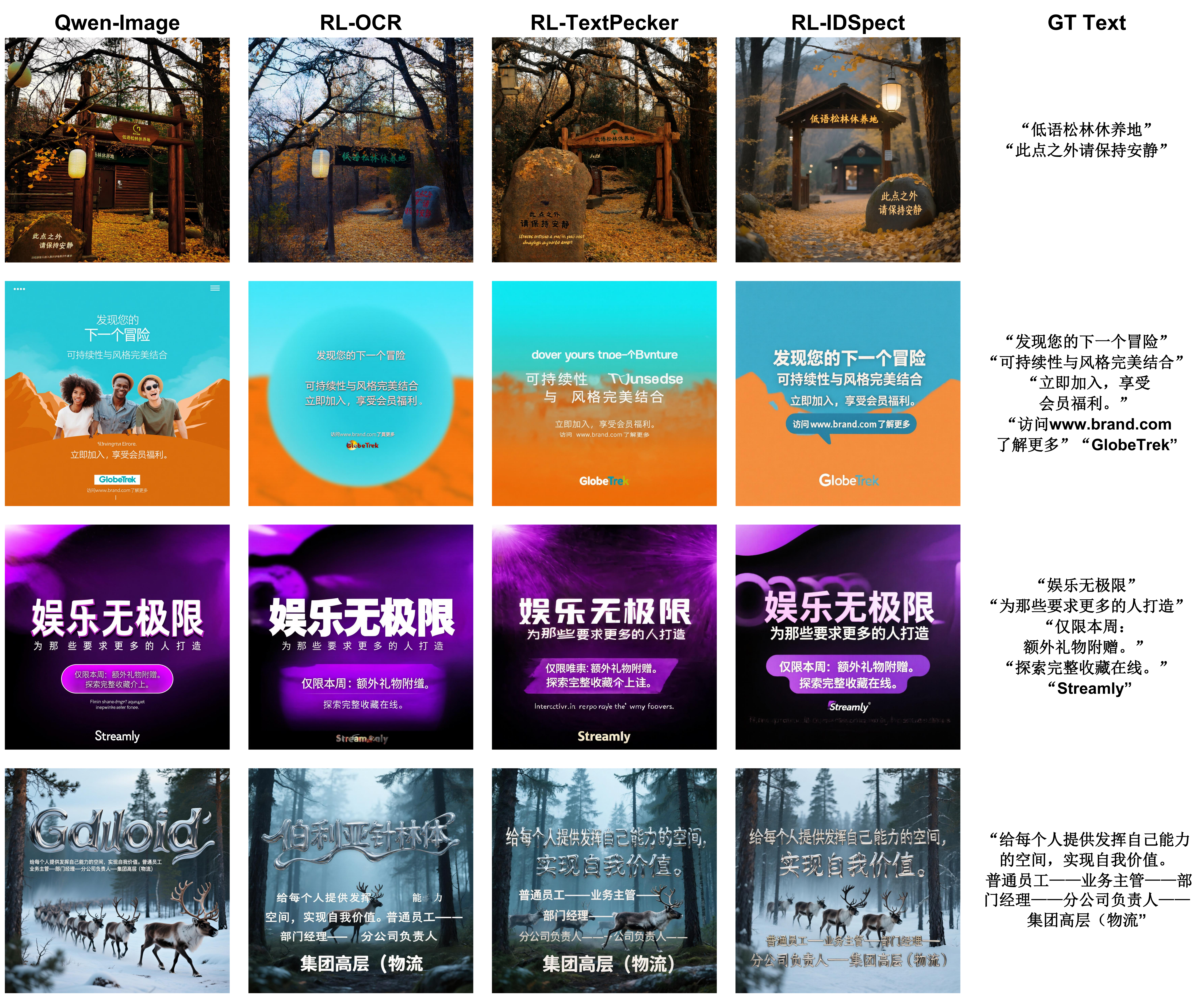}
  \caption{Qualitative comparison of Qwen-Image and OCR-, TextPecker-, and
  \method-guided GRPO on Chinese text-rendering prompts. The rightmost column
  lists the target text.}
  \label{fig:main_visual}
\end{figure}

\textbf{Qualitative results.}
Fig.~\ref{fig:main_visual} compares the frozen Qwen-Image model with OCR-, TextPecker-, and \method-guided GRPO on three prompts containing multiple Chinese text regions with different lengths and spatial arrangements. The frozen model frequently produces incomplete or corrupted characters, particularly in longer text lines. OCR-guided GRPO improves the overall readability of the generated text, while TextPecker-guided GRPO further improves character-level fidelity. In comparison, \method{} renders the requested main and secondary text more completely and better preserves the internal structure of individual Chinese characters. The improvement is especially apparent in longer lines on signs and posters, where structural errors can accumulate across characters and become difficult to capture with a coarse text-level reward. At the same time, \method{} maintains coherent scene layouts and the intended spatial organization of multiple text regions. These observations are consistent with the quantitative results and suggest that the IDS reward provides more detailed structural feedback during GRPO optimization.

\subsection{Ablation of IDS reward construction}
\label{sec:ablation}

\textbf{IDS reference.}
We first compare two types of IDS references. Per-crop consistency uses the IDS prediction of the OCR transcript recognized from each generated crop as the reference. This formulation measures whether the visual content is structurally consistent with the OCR result, but does not directly enforce agreement with the requested target text. In contrast, the other three variants use the canonical IDS decomposition of the target text as the reference, which directly connects the structural reward to the desired output. The semantic OCR term remains target-conditioned in all experiments. Per-crop consistency achieves the highest Qua. score of 0.980, showing that comparison against an OCR-derived structural representation provides a strong signal for local structural quality. However, its Sem. score is 0.901, which is lower than the 0.911 achieved by our target-based crop-wise alignment. This gap indicates that structural consistency with the OCR prediction does not necessarily guarantee alignment with the requested text, motivating the use of target IDS as the reference for structural supervision.

\textbf{Target-based comparison.}
We next compare three strategies for matching predicted IDS sequences against the target IDS representation. Direct concatenation compares crop-level predictions with the target IDS sequence according to detector order, making the reward sensitive to mismatches between detector order and target text order. Character-wise matching follows the Chinese matching strategy released with TextPecker~\cite{TextPecker}, which reduces dependence on crop ordering but treats characters independently and therefore discards useful inter-character ordering information. Our crop-wise alignment instead matches each detected text crop to a corresponding span of the target text while preserving the order of IDS tokens within each crop and enforcing unique target-token credit. As shown in Tab.~\ref{tab:ablation}, crop-wise alignment achieves the highest Sem. score of 0.911, improving over direct concatenation and character-wise matching by 0.025 and 0.012, respectively. Its Qua. score reaches 0.979, only 0.001 below per-crop consistency while providing substantially stronger semantic alignment. These results demonstrate that the effectiveness of the IDS reward depends not only on the structural representation itself but also on how the predicted structures are aligned with the target. Target-based supervision grounds the structural reward in the requested text, while crop-wise alignment preserves the correspondence between local text regions and their target spans.

\begin{table}[tbp]
\centering
\small
\caption{Ablation of IDS reward construction on GenTextEval. All variants use
the same balanced IDS recognizer, semantic OCR term, and training configuration.}
\label{tab:ablation}
\setlength{\tabcolsep}{9mm}
\begin{tabular}{llcc}
\toprule
IDS reference & Comparison & Qua. $\uparrow$ & Sem. $\uparrow$ \\
\midrule
OCR prediction & Per-crop consistency & \textbf{0.980} & 0.901 \\
Target text & Direct concatenation & 0.976 & 0.886 \\
Target text & Character-wise matching & 0.971 & 0.899 \\
Target text & \textbf{Crop-wise alignment (ours)} & 0.979 & \textbf{0.911} \\
\bottomrule
\end{tabular}
\end{table}

\section{Conclusion}
\label{sec:conclusion}

In this paper, we addressed a fundamental limitation of OCR-based rewards for Chinese visual text rendering: treating each ideograph as an atomic label provides insufficient credit for partial improvements to its internal structure. We introduced \textbf{\method{}}, a target-conditioned compositional reward that represents Chinese text with Ideographic Description Sequences and combines character-level semantic fidelity with fine-grained supervision over components and spatial operators. Experiments with GRPO post-training of Qwen-Image show that \method{} improves structural quality and semantic alignment on LongText-Bench and GenTextEval-Bench, outperforming general OCR-based rewards and providing stronger supervision than glyph-level anomaly-aware evaluation. These results demonstrate that decomposing ideographs into reusable components and explicit spatial relations offers a practical route to more faithful Chinese text generation. In future work, we plan to extend the framework to broader Chinese character coverage and multilingual scripts like Japanese and Korean.

\bibliographystyle{plainnat}
\bibliography{refs}

\end{document}